\PassOptionsToPackage{dvipsnames}{xcolor} 
\documentclass{article}

 \PassOptionsToPackage{numbers, compress, square}{natbib}
 
\usepackage[preprint]{nips_2018} 

\usepackage[utf8]{inputenc} 
\usepackage[T1]{fontenc}    
\usepackage{hyperref}       
\usepackage{url}            
\usepackage{booktabs}       
\usepackage{amsfonts}       
\usepackage{nicefrac}       
\usepackage{microtype}      

\title{\mbox{Phase transition in the knapsack problem}}

\author{Nitin Yadav \\The University of Melbourne \\ \texttt{nitin.yadav@unimelb.edu.au} \And Carsten Murawski \\The University of Melbourne \\ \texttt{carstenm@unimelb.edu.au} \AND Sebastian Sardina \\RMIT University \\ \texttt{sebastian.sardina@rmit.edu.au}  \And Peter Bossaerts \\The University of Melbourne \\ \texttt{peter.bossaerts@unimelb.edu.au}}
\usepackage[textsize=scriptsize]{todonotes}
\usepackage{booktabs}
\usepackage[final]{changes}

\usepackage{subcaption}
\usepackage{graphicx} 
  \setkeys{Gin}{width=\linewidth,totalheight=\textheight,keepaspectratio}
  \graphicspath{{graphics/}} 
\usepackage{amsmath}  
\usepackage{booktabs} 
\usepackage{units}    
\usepackage{multicol} 
\usepackage{fancyvrb} 
  \fvset{fontsize=\normalsize}
\usepackage{bm}

\usepackage{amssymb,amsmath,amsthm,amsfonts}
\usepackage{xspace}
\usepackage{pgf,tikz}
\usetikzlibrary{arrows,decorations,backgrounds,automata,shapes,patterns,petri,positioning}
\usepackage{tikz-qtree}
\usepackage{multirow}
\usepackage{color, colortbl}
\usepackage{float}
\usepackage{verbatim}
\usepackage{url}
\usepackage{stmaryrd}
\usepackage{latexsym}
\usepackage{algorithm}
\usepackage{algorithmic}
\usepackage{enumitem}
\usepackage{listings}
\usepackage{pgfplots}
\usepackage{graphicx}
\usepackage{stmaryrd}
\usepackage{epsfig}
\usepackage{epstopdf}
\usepackage{xspace}
\usepackage[all]{xy}
\usepackage{pifont}
\usepackage{ifthen}
\usepackage{fancyvrb}
\usepgfplotslibrary{colorbrewer}
\usepackage{rotating}

\def\qed{\hfill{\qedboxempty}      
  \ifdim\lastskip<\medskipamount \removelastskip\penalty55\medskip\fi}

\def\qedboxempty{\vbox{\hrule\hbox{\vrule\kern3pt
                 \vbox{\kern3pt\kern3pt}\kern3pt\vrule}\hrule}}

\def\qedfull{\hfill{\qedboxfull}   
  \ifdim\lastskip<\medskipamount \removelastskip\penalty55\medskip\fi}

\def\qedboxfull{\vrule height 4pt width 4pt depth 0pt}

\newcommand{\defterm}[1]{\mbox{\underline{\it\smash{#1}\vphantom{\lower.1ex\hbox{x}}}}}
\renewcommand{\defterm}[1]{\textbf{\emph{#1}}}

\newcommand{\set}[1]{\{#1\}}                      

\newcommand{\tup}[1]{\langle #1\rangle}            
\newcommand{\tuple}[1]{\tup{#1}}            

\definecolor{darkgrey}{RGB}{45,45,45}
\tikzstyle{every initial by arrow}=[initial text=]
\tikzstyle{every edge}=[draw=Gray]
\tikzstyle{every state}=[fill=none,draw=Gray,text=darkgrey,inner sep=0pt,minimum
size=10mm]
\tikzstyle{every picture}=[->,>=stealth',shorten >=1pt,auto,node distance=2.5cm,
semithick]
\tikzstyle{sim}=[->,dotted]
\tikzstyle{estate}=[state,circle split,text centered,inner sep=1pt]

\tikzset{ActionStyle/.style = {font=\small}}

\newcommand{\ept}{\ensuremath{\mathbb{E}}}

\newcommand{\lbl}[1]{\footnotesize{#1}}

\renewenvironment{quote}{%
   \list{}{%
     \leftmargin0.5cm   
     \rightmargin\leftmargin
   }
   \item\relax
}
{\endlist}
\begin{document}

\maketitle

\begin{abstract}
We examine the phase transition phenomenon for the Knapsack problem from both a computational and a human perspective.
We first provide, via an empirical and a theoretical analysis, a characterization of the phenomenon in terms of two instance properties; \emph{normalised capacity} and \emph{normalised profit}. 
%
Then, we show evidence that average time spent by human decision makers in solving an instance peaks near the phase transition. 
Given the ubiquity of the Knapsack problem in every-day life, a better understanding of its structure can improve our understanding not only of computational techniques but also of human behavior, including the use and development of heuristics and occurrence of biases. 
%
\end{abstract}
  
\section{Introduction}\label{sec:introduction}

The Knapsack problem (KP) is an important and relevant problem from both a \emph{computational} and a \emph{human reasoning} perspective: it is a classical NP-complete problem that is ubiquitous in every-day human tasks at many different levels of 
cognition~\cite{Torralva:2013ADHD,MurawskiBossaerts:SR16}. 
In this work, we study---theoretically and empirically---its ``hardness" properties and whether these coincide for computers and humans.

A computational complexity analysis of the problem (known to be in the NP-complete complexity class) provides too conservative a view, in that it is based on worst-case behaviour that may not coincide with `typical' case instances. Thus, we aim to analyse the relation between instance properties and instance complexity of the problem.
To do so, we provide a fine-grained analysis of the \emph{phase transition phenomenon} for the Knapsack problem, a sudden transition from solubility to insolubility within a narrow range of instance parameters. 
Studying phase transitions in solubility of instances is considered to be one promising approach to gain a better, more meaningful, understanding of instance complexity~\cite{HoggHubermanWilliams:BOOK96}.
Indeed, such phenomenon has been documented for several other NP-complete problems, such as SAT~\cite{MitchellSelmanLevesque:AAAI92-PhaseTransition}, Hamiltonian Circuit~\cite{GareyJhonson:NPBook}, and Integer Partitioning~\cite{GareyJhonsonNPResults:1979}, as well as for many physical systems, with complexity peaking at the phase boundary~\cite{Monasson.etal:NATURE92-PhaseTransition}.

More concretely, we first show that the \emph{decision variant} of the (0-1) Knapsack problem displays a clear phase transition in solubility for random instances, and we identify its region in terms of two easily observable instance properties: \emph{instance} capacity and profit.
We provide a theoretical analysis that utilizes a fast greedy-type estimation rule to provide a lower bound of the true chance that a given instance admits a solution, and confirm the phenomena via experimental analysis as well.
We also demonstrate that, similar to other NP-complete problems, the computationally hard instances happen to lie on the boundary of the phase transition.
Importantly, in our analysis, we show an interesting ``convexity'' property of the phase transition region.

Secondly, we provide empirical evidence that indicates that instances that are challenging for humans also occur near the identified phase transition.
This finding is significant in that understanding complexity of instances may be relevant for many aspects of human behavior~{\cite{MurawskiBossaerts:SR16}}.
Indeed, the KP occurs at many different levels of cognition, including attention~\cite{Torralva:2013ADHD}, intellectual discovery~\cite{Meloso:2009go} and investment decisions~\cite{Mansini:1999PortOpt}.
~\citet{MurawskiBossaerts:SR16} have recently demonstrated that human performance in the KP \added{(optimisation variant)} is correlated with instance complexity. The authors argue, there, that their findings can explain several important anomalies in human behaviour, such as choice overload~\cite{Alvin:1970futureshock} and the use of the availability heuristic~\cite{TverskyKahneman:1975judgment}.
%
However, their work uses an algorithmic specific property, namely the ``Sahni-k'' \cite{sahni-k:1997} metric, to provide an \emph{ex-post} complexity measure of KP instances.

We note that there exists work on instance difficulty for the KP (e.g.,~\cite{Pisinger:JCOR05,OptInstanceComplexity:Smith12,Kellerer.etal:BOOK04-Knapsack}). 
However, such works focused on studying the performance of \replaced{specific}{particular (approximate)} algorithms \replaced{tailored}{developed specifically} for the optimisation variant of the \replaced{KP}{knapsack problem}. Particularly, they use the correlation between weights and values of items to identify if the instances were hard or easy for specific algorithms.
Importantly, such studies identify no phase transition or its relation with computational and human reasoning complexity, \added{which is} the main objective of our work.

\section{The Knapsack Problem}\label{sec:knapsack-problem}
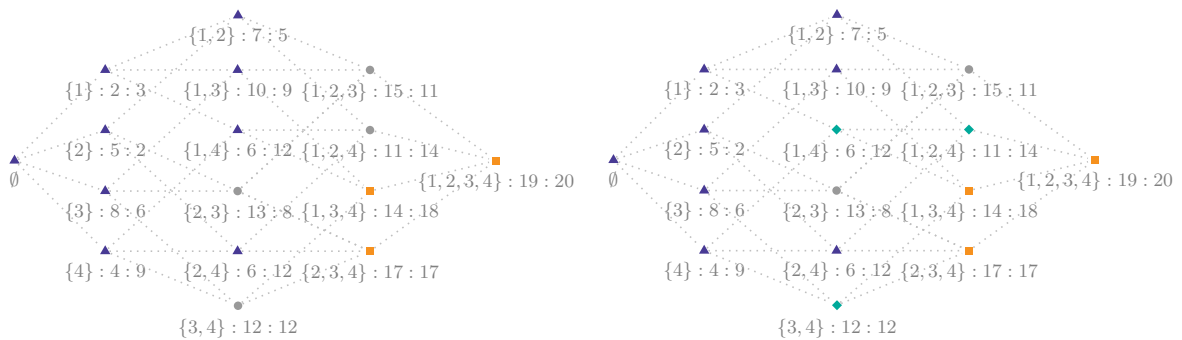
\begin{figure*}[!t]
\begin{subfigure}{0.45\textwidth}

\begin{tikzpicture}[node distance=0.8cm, on grid,-,scale=0.7, every node/.style={scale=0.8}]
  \tikzstyle{node}=[circle,fill=gray!80,minimum size=4pt, inner sep=1pt]
  \tikzstyle{cap-feasible}=[regular polygon,regular polygon sides=3,fill=BlueViolet,minimum size=4pt, inner sep=1pt]
  \tikzstyle{val-feasible}=[regular polygon,regular polygon sides=4,fill=BurntOrange,minimum size=5pt, inner sep=1pt]
  \tikzstyle{every label}=[gray]
  \tikzstyle{every edge}=[draw=gray!50]
  \begin{scope} 
    \node (0) [cap-feasible,label={below:\lbl{$\emptyset$}}] {};
    \node (2) [cap-feasible, xshift=0.8cm, yshift=-0.2cm, above right = of 0, label={below:\lbl{$\set{2}:5:2$}}] {};
    \node (3) [cap-feasible, xshift=0.8cm, yshift=0.2cm, below right = of 0,label={below:\lbl{$\set{3}:8:6$}}] {};
    \node (1) [cap-feasible, above = of 2,label={below:\lbl{$\set{1}:2:3$}}] {};
    \node (4) [cap-feasible, below = of 3,label={below:\lbl{$\set{4}:4:9$}}] {};
    \node (12) [cap-feasible, xshift=1.5cm, yshift=0.2cm, above right = of 1,label={below:\lbl{$\set{1,2}:7:5$}}] {};
    \node (13) [cap-feasible, xshift=1.2cm, right = of 1,label={below:\lbl{$\set{1,3}:10:9$}}] {};
    \node (14) [cap-feasible, xshift=1.2cm, right = of 2,label={below:\lbl{$\set{1,4}:6:12$}}] {};
    \node (23) [node, xshift=1.2cm, right = of 3,label={below:\lbl{$\set{2,3}:13:8$}}] {};
    \node (24) [cap-feasible, xshift=1.2cm, right = of 4,label={below:\lbl{$\set{2,4}:6:12$}}] {};
    \node (34) [node, xshift=1.5cm, yshift=-0.2cm, below right = of 4,label={below:\lbl{$\set{3,4}:12:12$}}] {};
    \node (123) [node, xshift=1.2cm, right = of 13,label={below:\lbl{$\set{1,2,3}:15:11$}}] {};
    \node (124) [node, xshift=1.2cm, right = of 14,label={below:\lbl{$\set{1,2,4}:11:14$}}] {};
    \node (134) [val-feasible, xshift=1.2cm, right = of 23,label={below:\lbl{$\set{1,3,4}:14:18$}}] {};
    \node (234) [val-feasible, xshift=1.2cm, right = of 24,label={below:\lbl{$\set{2,3,4}:17:17$}}] {};
    \node (1234) [val-feasible, xshift=7cm, right = of 0,label={below:\lbl{$\set{1,2,3,4}:19:20$}}] {};

    \path (0) edge [dotted] (1);
    \path (0) edge [dotted] (2);
    \path (0) edge [dotted] (3);
    \path (0) edge [dotted] (4);

    \path (1) edge [dotted] (12);
    \path (1) edge [dotted] (13);
    \path (1) edge [dotted] (14);
    \path (2) edge [dotted] (23);
    \path (2) edge [dotted] (12);
    \path (2) edge [dotted] (24);
    \path (3) edge [dotted] (13);
    \path (3) edge [dotted] (23);
    \path (3) edge [dotted] (34);
    \path (4) edge [dotted] (14);
    \path (4) edge [dotted] (24);
    \path (4) edge [dotted] (34);

    \path (12) edge [dotted] (123);
    \path (12) edge [dotted] (124);
    \path (13) edge [dotted] (123);
    \path (13) edge [dotted] (134);
    \path (14) edge [dotted] (124);
    \path (14) edge [dotted] (134);
    \path (23) edge [dotted] (123);
    \path (23) edge [dotted] (234);
    \path (34) edge [dotted] (134);
    \path (34) edge [dotted] (234);
    \path (23) edge [dotted] (234);
    \path (23) edge [dotted] (123);
    \path (24) edge [dotted] (124);
    \path (24) edge [dotted] (234);

    \path (123) edge [dotted] (1234);
    \path (124) edge [dotted] (1234);
    \path (134) edge [dotted] (1234);
    \path (234) edge [dotted] (1234);
  \end{scope}

\end{tikzpicture}
  \caption{\small{Weight and profit constraints of $10$ and $15$, resp.}}
  \label{fig:KP-example-nosol}
\end{subfigure}\hspace{0.12\textwidth}%
\begin{subfigure}{0.45\textwidth}

\begin{tikzpicture}[node distance=0.8cm, on grid,-,scale=0.7, every node/.style={scale=0.8}]
  \tikzstyle{node}=[circle,fill=gray!80,minimum size=4pt, inner sep=1pt]
  \tikzstyle{cap-feasible}=[regular polygon,regular polygon sides=3,fill=BlueViolet,minimum size=4pt, inner sep=1pt]
  \tikzstyle{val-feasible}=[regular polygon,regular polygon sides=4,fill=BurntOrange,minimum size=5pt, inner sep=1pt]
  \tikzstyle{feasible}=[diamond,fill=JungleGreen,minimum size=5pt, inner sep=1pt]
  \tikzstyle{every label}=[gray]
  \tikzstyle{every edge}=[draw=gray!50]
  \begin{scope}
    \node (0) [cap-feasible,label={below:\lbl{$\emptyset$}}] {};
    \node (2) [cap-feasible, xshift=0.8cm, yshift=-0.2cm, above right = of 0, label={below:\lbl{$\set{2}:5:2$}}] {};
    \node (3) [cap-feasible, xshift=0.8cm, yshift=0.2cm, below right = of 0,label={below:\lbl{$\set{3}:8:6$}}] {};
    \node (1) [cap-feasible, above = of 2,label={below:\lbl{$\set{1}:2:3$}}] {};
    \node (4) [cap-feasible, below = of 3,label={below:\lbl{$\set{4}:4:9$}}] {};
    \node (12) [cap-feasible, xshift=1.5cm, yshift=0.2cm, above right = of 1,label={below:\lbl{$\set{1,2}:7:5$}}] {};
    \node (13) [cap-feasible, xshift=1.2cm, right = of 1,label={below:\lbl{$\set{1,3}:10:9$}}] {};
    \node (14) [feasible, xshift=1.2cm, right = of 2,label={below:\lbl{$\set{1,4}:6:12$}}] {};
    \node (23) [node, xshift=1.2cm, right = of 3,label={below:\lbl{$\set{2,3}:13:8$}}] {};
    \node (24) [cap-feasible, xshift=1.2cm, right = of 4,label={below:\lbl{$\set{2,4}:6:12$}}] {};
    \node (34) [feasible, xshift=1.5cm, yshift=-0.2cm, below right = of 4,label={below:\lbl{$\set{3,4}:12:12$}}] {};
    \node (123) [node, xshift=1.2cm, right = of 13,label={below:\lbl{$\set{1,2,3}:15:11$}}] {};
    \node (124) [feasible, xshift=1.2cm, right = of 14,label={below:\lbl{$\set{1,2,4}:11:14$}}] {};
    \node (134) [val-feasible, xshift=1.2cm, right = of 23,label={below:\lbl{$\set{1,3,4}:14:18$}}] {};
    \node (234) [val-feasible, xshift=1.2cm, right = of 24,label={below:\lbl{$\set{2,3,4}:17:17$}}] {};
    \node (1234) [val-feasible, xshift=7cm, right = of 0,label={below:\lbl{$\set{1,2,3,4}:19:20$}}] {};

    \path (0) edge [dotted] (1);
    \path (0) edge [dotted] (2);
    \path (0) edge [dotted] (3);
    \path (0) edge [dotted] (4);

    \path (1) edge [dotted] (12);
    \path (1) edge [dotted] (13);
    \path (1) edge [dotted] (14);
    \path (2) edge [dotted] (23);
    \path (2) edge [dotted] (12);
    \path (2) edge [dotted] (24);
    \path (3) edge [dotted] (13);
    \path (3) edge [dotted] (23);
    \path (3) edge [dotted] (34);
    \path (4) edge [dotted] (14);
    \path (4) edge [dotted] (24);
    \path (4) edge [dotted] (34);

    \path (12) edge [dotted] (123);
    \path (12) edge [dotted] (124);
    \path (13) edge [dotted] (123);
    \path (13) edge [dotted] (134);
    \path (14) edge [dotted] (124);
    \path (14) edge [dotted] (134);
    \path (23) edge [dotted] (123);
    \path (23) edge [dotted] (234);
    \path (34) edge [dotted] (134);
    \path (34) edge [dotted] (234);
    \path (23) edge [dotted] (234);
    \path (23) edge [dotted] (123);
    \path (24) edge [dotted] (124);
    \path (24) edge [dotted] (234);

    \path (123) edge [dotted] (1234);
    \path (124) edge [dotted] (1234);
    \path (134) edge [dotted] (1234);
    \path (234) edge [dotted] (1234);
  \end{scope}

\end{tikzpicture}
  \caption{\small{Weight and profit constraints of $12$ and $12$, resp.}}
  \label{fig:KP-example-sol}
\end{subfigure}
\caption{Search space of a knapsack instance with 4 items with weights $\set{2,5,8,4}$ and values $\set{3,2,6,9}$. Nodes that satisfy the weight constraint are indicated by a triangle, nodes that satisfy the profit constraint are indicated by a square, and nodes that satisfy both constrains are indicated by a diamond.}
\label{fig:KP-example}
\end{figure*}
The 0-1 knapsack problem is a combinatorial optimisation problem with the goal of finding the subset of a set of items with given values and weights that maximises total profit subject to a capacity (weight) constraint. 
%
The knapsack problem is NP-hard. The number of subsets that can be formed from the $n$ available items increases exponentially in $n$ ($2^n$). However, as is the case for some NP-hard problems \cite{MitchellSelmanLevesque:AAAI92-PhaseTransition}, many instances can be solved in polynomial time, even if $n$ is large.
%
It has been shown for some NP-complete problems that hardness appears to be closely related to phase transitions in solvability of instances \cite{Cheeseman:PT1991,natureOfComputation}, which typically occurs within a narrow range of instance properties. It is an open question whether this is also the case for the knapsack problem.

To address this question, we will consider here the \emph{decision variant} of the knapsack problem: \emph{Does there exists a subset of items from the set of available items with total profit at least as high as a given profit constraint, and total weight at most as high as a given capacity constraint?} Formally,
given a set of items $I = \set{1,\ldots,n}$ with weights $\tuple{\hat{w}_1,\ldots,\hat{w}_n}$ and values $\tuple{\hat{v}_1,\ldots,\hat{v}_n}$, where each $\hat{w}_i$ and $\hat{v}_i$ is a positive integer, and two positive integers $\hat{c}$ and $\hat{p}$ denoting the capacity and profit constraint (of the knapsack) \emph{decide whether there exists} a knapsack set $S \subseteq I$ such that:
\begin{itemize}
  \item $\sum\limits_{i \in S}\hat{w}_i \leq \hat{c}$, the weight of the knapsack is less than or equal to the capacity constraint; and

  \item $\sum\limits_{i \in S}\hat{v}_i \geq \hat{p}$, the value of the knapsack is greater than or equal to the profit constraint.
\end{itemize}
The decision variant of the knapsack problem is closely related to the optimisation variant; the latter can be solved by iteratively solving the former while incrementing the profit constraint.

The knapsack problem is, basically, a constrained satisfaction problem with two opposing constraints. Increasing the profit level requires adding items to the knapsack, while decreasing the weight requires removing items from the knapsack.

We use a small example to highlight this tension between the two constraints.
Consider an instance of the knapsack problem consisting of $4$ items with weights $\tuple{2,5,8,4}$  and profits $\tuple{3,2,6,9}$.
The graphs shown in Figure~\ref{fig:KP-example} depict the search space for this instance (for two different capacity and profit constraints).
Each node in the graph is a knapsack configuration and the edges link the knapsack configurations that are reachable by addition or deletion of a single item. For example, the node $\set{1,3}:10:9$ represents a knapsack containing two items $1$ and $3$, with a total weight of $10$ and total profit of $9$. This node can be reached in 4 ways (there are 4 edges linking this node), one of which is to add item $3$ to a knapsack that already contains item $1$.
The layout of the graph is such that, as we traverse from left to right, the number of items in the knapsack increases. The vertical alignment is representative of the number of subsets of fixed cardinality (i.e. $\binom{n}{k}$). We use the empty knapsack (left most node) as the initial node of the search space.

Suppose the capacity and the profit constraints for this instance are $10$ and $15$, respectively. This is the case of Figure~\ref{fig:KP-example-nosol}. Triangle nodes represent configurations that satisfy the capacity constraint, whereas square nodes denote configurations that satisfy the profit constraint, and diamond nodes satisfy both the constraints (i.e, these are solution nodes).
As is evident from the graph, the nodes that satisfy the capacity constraint are towards the left and the nodes that satisfy the profit constraint are towards the right. Note that, in general, as the the capacity constraint is relaxed, the number of nodes that satisfy the constraint increases proportional to $\sum_{i=0}^{k}\binom{n}{i}$, where $n$ is the total number of items, and $k$ is proportional to how ``relaxed'' the constraint is. Similar reasoning applies to the profit constraint.

Informally, when both constraints are too ``tight,'' no solution nodes will exist, as in the case of Figure~\ref{fig:KP-example-nosol}.
When the constraints are relaxed, the number of solution nodes will tend to increase and the answer to the decision problem of the instance will be yes. For example, Figure~\ref{fig:KP-example-sol} shows the search space when both capacity and profit constraint are $12$.

In the following, we will investigate whether solvability of instances of the knapsack problem exhibit a phase transition, that is, whether the probability that both constraints can be satisfied changes precipitously from near $1$ to near $0$.


\section{Probabilistic analysis}\label{sec:formal}

Here we characterize a phase transition phenomenon for random instances of the Knapsack problem.
Importantly, we also show a ``convexity'' property of the phase transition region. 
Our analysis utilizes a fast greedy-type estimation rule to provide a lower bound of the true chance that a given instance has a solution.

\paragraph{Existence of Phase Transitions.}
We consider a distribution of Knapsack instances with varying item weights and values, keeping the number of items fixed at $n$.
Intuitively, a phase transition emerges around critical values of capacity and profit for which the probability that a random instance has a solution changes from zero to one.

To achieve meaningful comparisons across instances, we normalize capacity and profit. 
Given a tuple of positive integers $\tuple{\hat{x}_1,\ldots,\hat{x}_n}$ and a real number $\hat{y} \in [0,\sum_{i=1}^{n}\hat{x}_i]$, we shall use the function $\sigma(\hat{y}, \tuple{\hat{x}_1,\ldots,\hat{x}_n}) = \frac{\hat{y}}{\sum_{i=1}^{n}\hat{x}_i}$ as the \emph{normalized value} of $\hat{y}$ with respect to $\tuple{\hat{x}_1,\ldots,\hat{x}_n}$.
Note that the range of the function $\sigma$ is $[0,1]$.

So, given a knapsack instance with items $I = \set{1,\ldots,n}$ having weights $\tuple{\hat{w}_1,\ldots,\hat{w}_n}$, values $\tuple{\hat{v}_1,\ldots,\hat{v}_n}$, capacity constraint $\hat{c}$ and profit constraint $\hat{p}$, we denote the \emph{normalized capacity constraint} by $c=\sigma(\hat{c}, \tuple{\hat{w}_1,\ldots,\hat{w}_n})$ and the \emph{normalized profit constraint} by $p=\sigma(\hat{p}, \tuple{\hat{v}_1,\ldots,\hat{v}_n})$.
A set of items $S\subseteq I$ is a \emph{solution} to an instance when \emph{(a)} the normalized sum of item weights in $S$ is at most the normalized capacity, that is, $\sigma(\sum_{i\in S}\hat{w}_i, \tuple{\hat{w}_1,\ldots,\hat{w}_n}) \leq c$; and \emph{(b)} the normalized sum of item values in $S$ is at least the normalized profit, that is, $\sigma(\sum_{i\in S}\hat{v}_i, \tuple{\hat{v}_1, \ldots, \hat{v}_n}) \geq p$.
\emph{In the following we use this normalized version as our canonical definition of a knapsack instance} and  look at phase transition across two measures: \emph{(i)} the space $(c,p)$ of normalized capacity and profit values; and \emph{(ii)} the space $r=c/p$ of the ratio between normalized capacity and profit.\footnote{We denote the normalized value of a parameter $\hat{x}$ by $x$.}


We do not impose any restrictions on the distribution from which values and weights are drawn for the current set of items, except that they are continuous, identical, and independent. An example would be gamma-distributed values and weights, in which case the normalized values and weights are Dirichlet~\cite{MultiVariateStats:Bilodeau1999}.

For capacity constraint, we shall consider $c$ (normalized) values within $[0,1]$.
For the profit constraint, in turn, we look for a (normalized) $p$ that can always be met by any sampled instance. Note that, the ratio $c/p$ does not exist for $p = 0$.
So we will consider values of $p$ within $[p_{\min},1]$, where $p_{\min} = 1/n$. Since an item with a normalized value at least $1/n$ is guaranteed to always exist in any sampled instance (because of the restriction that all values are drawn from identical distributions and that normalized item values add up to $1$).
As a result, the ratio $r = c/p \in [0,n]$.
 

%
We now aim to better understand the boundaries of the phase transition region. We do so by studying the event $E(c,p)$ (i.e., analysing its chances): collection of weights and values for all $n$ items for which the resulting KP instances (with capacity $c$ and profit $p$) admit a solution.

%

\paragraph{Lower bounds for \bm{$P[E(c,p)]$}.} We first consider a weak bound.
 Given a KP instance, we order the items arbitrarily, thus creating arbitrary sequences of values and weights that (after normalization) add up to $1$.
We then consider knapsacks with increasing number of items $s \geq 1$, which we fill in the order of the sequence. 
For the knapsack with the first $s$ items, we determine whether the capacity and the profit constraint are met. The process is similar to executing an incomplete algorithm that at each step adds the ``next" item to the knapsack and checks if the capacity and profit constraints are satisfied. In the worst case, such algorithm would execute $n$  such steps.

We consider the event $E^l(c,p)$ as a collection of weights and values assignment for the items together with \emph{an ordering on those items} such that for each assignment there is only one ordering in the event set, and such that the capacity and profit constraints $c$ and $p$, resp., are met by taking the first $s \geq 1$ items in the corresponding ordering. 
It follows then that $P[E^l(c,p)]$ provides {\em lower bounds} for $P[E(c,p)]$ since, under $E^l$, we only consider one possible ordering for each instance (a re-ordering of the items may make the capacity and profit constraints hold when the original, random ordering, did not).
Formally, $P[E^l(c,p)] \leq P[E(c,p)]$, for all $c$ and $p$.

To get a closed expression for $P[E^l(c,p)]$ we sum the probability of a solution over the possible size of knapsacks (i.e., from 0 to $n$).  The probability that a knapsack of size $k$ is a solution is equal to the product of two probabilities: (i) the probability that the capacity constraint is satisfied \emph{exactly} at $k$ (a partition on the capacity constraint) and, (ii) the probability that the profit constraint is satisfied.
Now, one can write $P[E^l(c,p)]$ more explicitly as follows (where the constraint $\sum_{i=1}^{n+1} w_i > c$ is assumed to be always true):
\vspace{-0.2cm}{
\small
\begin{equation}\label{eq:prob_lb}
	P[E^l(c,p)] = \sum_{s=0}^{n} f(s)\times G(s),
\end{equation}
}
where:
\begin{itemize}
  \item $f(s) =  P[ \sum_{i=1}^s w_i \leq c \wedge  \sum_{i=1}^{s+1} w_i > c ]$ denotes the probability (density) that the knapsack reaches the (max feasible) capacity at $s$; and
  
  \item $G(s) =  P[ \sum_{i=1}^s v_i \geq p ]$ denotes the \emph{cumulative} probability that the knapsack reaches the profit constraint with the first $s$ items. 
\end{itemize}

Integrating Equation~\ref{eq:prob_lb}, we then obtain:
\[
 P[E^l(c,p)] =1 - \sum_0^n F(s) g(s),
\]
where $F(s) = \sum_0^s f(s)$ and $g(s)= G(s) - G(s-1)$.

\smallskip
We now consider a \emph{stronger} bound by re-arranging the sequence of items so that they are in \emph{ascending order} of weights. We refer to this strategy as \emph{weight-greedy}, which does not affect $G(s)$ (or $g(s)$) since values are drawn independently. 
However, $\tilde{F}(s)$, the (cumulative) probability that the knapsack reaches capacity at $t \leq s$, is now smaller than $F(s)$, with strict inequality for some $s$. Indeed, if at $s$ items, capacity is reached with the unordered sequence, there is a positive chance that items in position $t$, $t > s$, are smaller (since the weights have to add up to 1), and hence, these would come earlier in the ordered sequence, implying that capacity is reached only at a larger $s$. Let $\tilde{f}(s) = \tilde{F}(s) - \tilde{F}(s-1)$.

Let $E^L(c,p)$ denote the event that a knapsack instance has a solution at $c$ and $p$ when the items are re-ordered according to increasing weight. Then, following a similar derivation to Equation~\ref{eq:prob_lb} we get:
\begin{eqnarray}
P[E^L(c,p)]  =  \sum_0^n \tilde{f}(s) G(s) =  1 - \sum_0^n \tilde{F}(s) g(s). \nonumber
\end{eqnarray} 

Since $\tilde{F}(s) < F(s)$, we have that $P[E^L(c,p)]$ is a better lower bound for $P[E(c,p)]$ (i.e., $P[E^L(c,p)] > P[E^l(c,p)]$ and $P[E^L(c,p)] \leq  P[E(c,p)]$).

We now show that \emph{location of the phase transition region is above the 45 degree line in $(c,p)$ space (or below $1$ in the $r$ space); and it exhibits a ``convexity'' shape}.
If $\tilde{F}$ and $g$ are non-negatively correlated as a function of $s$, we can then formally express the location of the isoquant where the lower bound equals $1$ as follows:
{\small
\begin{align*}
\sum_0^n \tilde{F}(s) g(s) \geq n \left( \frac{1}{n} \sum_0^n \tilde{F}(s) \frac{1}{n} \sum_0^n g(s) \right) > 1/n.
\end{align*}
}
Hence, $P[E^L(c,p)] = 1 - \sum_0^n \tilde{F}(s) g(s) < 1 - 1/n,$ or $P[E^L(c,p)] < 1$ for $c, p$ where $\tilde{F}$ and $g$ are non-negatively correlated.  
%
Non-negative correlation between $\tilde{F}$ and $g$ is most likely to occur for $c<p$:  $\tilde{F}$ is monotone increasing towards $1$ which it reaches for low $s = s^*$ ($s^*$ is the value at which knapsack is filled to capacity), after which it is flat; $g(s)$ is increasing in $s$ and will only decrease beyond $s^*$ since $p > c$. 
Note that, even if $c < p$, for $c$ close to $p$, and for small values of $c$, $\tilde{F}$ reaches 1 only at higher $s^*$ since the items are ordered by weight. At the same time, the profit constraint may be satisfied for low values of $s$, and $g(s)$ peaks before reaching $s^*$. 
The correlation constraints holds for most of the $(c,p)$ space when $c$ is sufficiently high, which implies that the lower bound of the contour of probabilities where $P[E^L(c,p)] = 1$ should be below the 45 degree line for high $c$. In $r$ space, the phase transition therefore stops at a value $r_1$ which is close to or below 1. 


\paragraph{Convexity.}
Now consider the case where $c = p$ (i.e., $r = 1$). Notice that, by symmetry (weights and values have the same distribution):
\[
	P[E^l(c,c)]  = \sum_0^n f(s) F(s)  =  	1 - \sum_0^n F(s) f(s),
\]
so $P[E^l(c,c)] = 1/2.$ 
Hence, it follows that
$
1/2 < P[E^L(c,c)] < P[E(c,c)] (= P[E(1)]).
$
That is, \emph{when $c = p$ (i.e., $r=1$), the chances that a solution exists is strictly larger than $1/2$}. 
As a result, \emph{maximum uncertainty (entropy) about the existence of a solution occurs in the region $c< p$, i.e., above the diagonal in $(c,p)$ space}. We refer to this as \emph{convexity of the region of phase transition}.


%
%
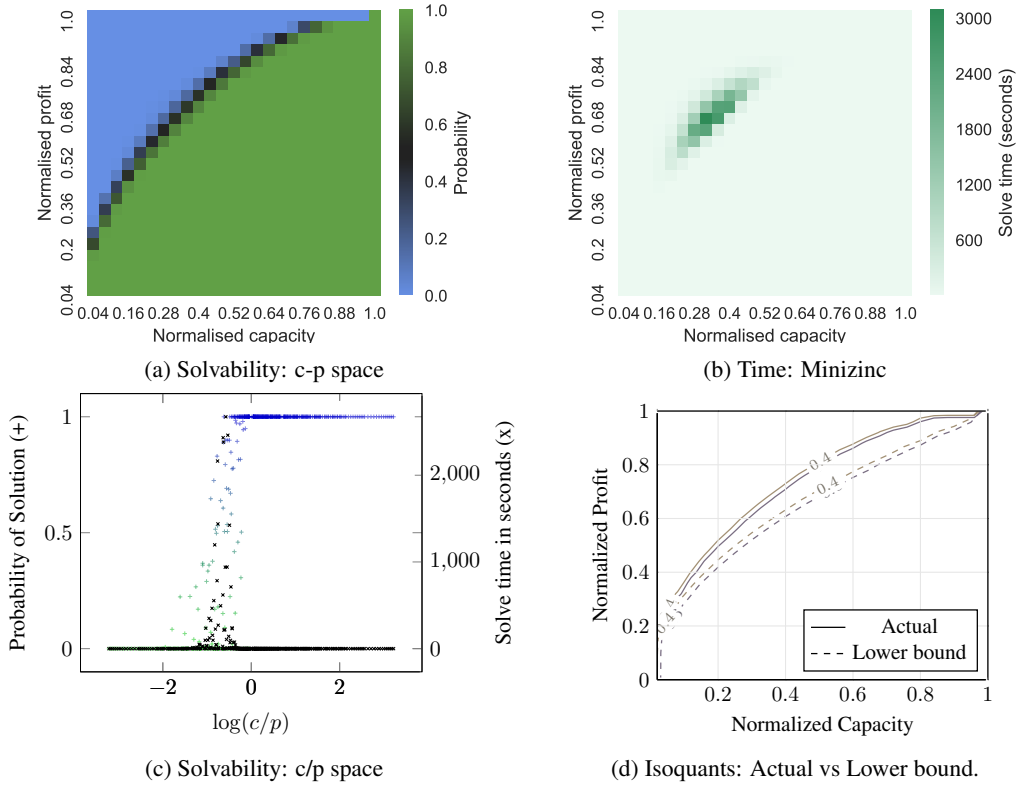
\begin{figure*}[!t]
  \hspace{-0.4cm}
\begin{tabular}{ccc}
\begin{subfigure}[b]{0.45\textwidth}
    \includegraphics[width=\textwidth]{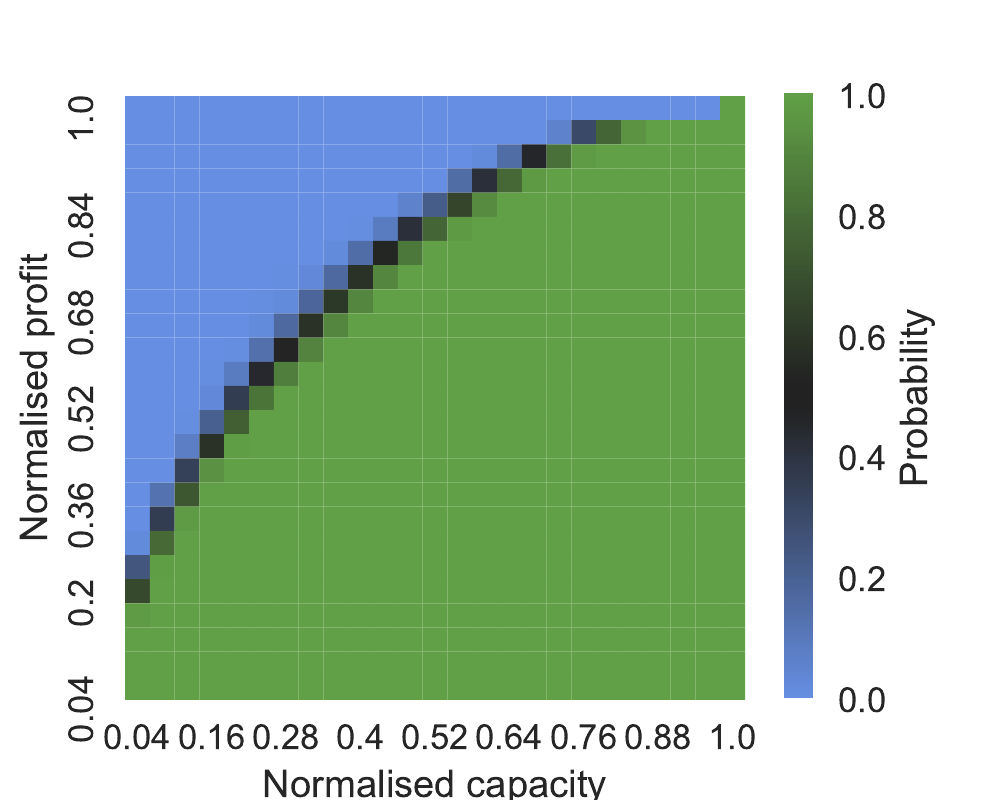}
    \caption{\small{Solvability: c-p space}}\label{fig:solve:c-p}
\end{subfigure}
&
\begin{subfigure}[b]{0.45\textwidth}
    \includegraphics[width=\textwidth]{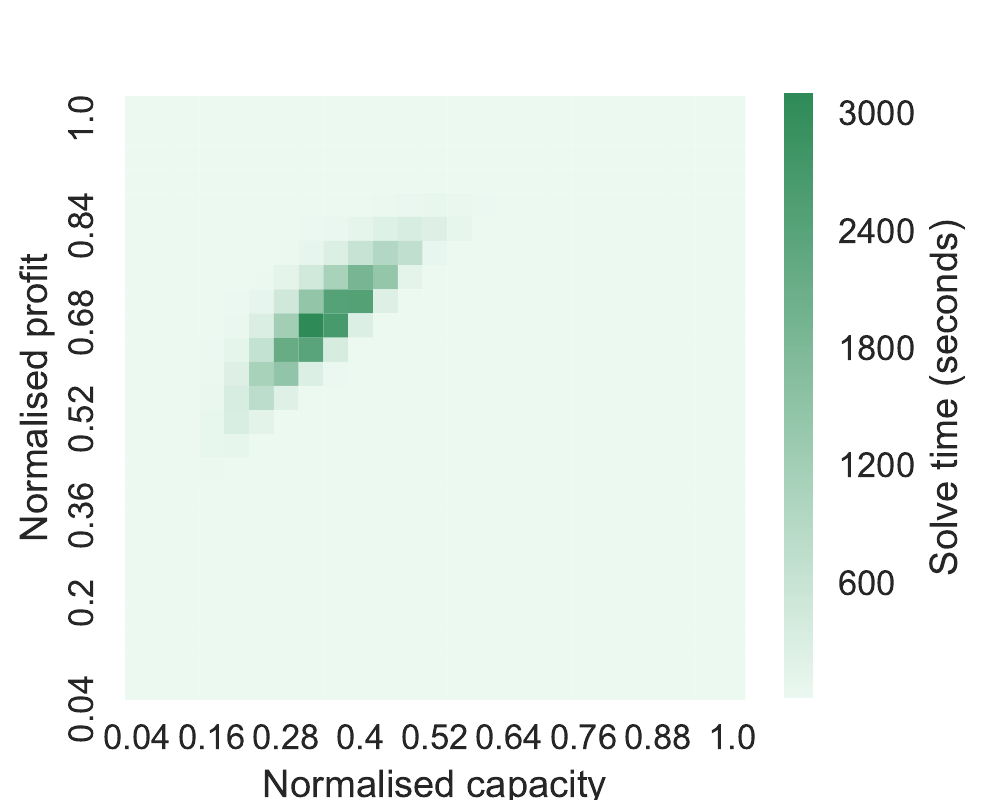}
     \caption{\small{Time: Minizinc}}\label{fig:stime-mzn}
\end{subfigure}
\\
\begin{subfigure}[b]{0.5\textwidth}
	\resizebox{\textwidth}{!}{

\begin{tikzpicture}
\tikzstyle{every picture}=[-]
	\begin{axis}[
		width=7cm,
		grid=minor,
		xlabel={$\log(c/p)$}, 
		ylabel={Probability of Solution (+)},
    	label style={sloped},
    	height=6cm,
 	    width=7cm,
 	    ylabel near ticks
    ]
	
	\addplot[
		scatter=true, 
		only marks, 
		mark=+, 
		mark size=1pt, 
		red,
		colormap={redyellow}{ rgb255(0cm)=(0,255,0);rgb255(1cm)=(0,0,255);},
		fill opacity=0.5,
		draw opacity=0.5
	] table[x=x, y=solvability, col sep=comma] {mzn-50-solvability.csv};
	\end{axis}

	\begin{axis}[
		width=7cm,
		grid=minor,
		ylabel={Solve time in seconds (x)},
		axis y line*=right,
		ylabel near ticks, yticklabel pos=right,
    	label style={sloped},
    	height=6cm,
 	    width=7cm,
 	    ylabel near ticks
    ]
	\addplot[
		scatter, 
		only marks, 
		mark=x, 
		mark size=1pt, 
	blue,
	colormap={redyellow}{ rgb255(0cm)=(0,0,0);rgb255(1cm)=(0,0,0);}
	] table[x=x, y=time, col sep=comma] {mzn-50-time.csv};
	\end{axis}
\end{tikzpicture}}
     \caption{\small{Solvability: c/p space}}\label{fig:solve:ratio}
\end{subfigure}
&
\begin{subfigure}[b]{0.4\textwidth}
	\resizebox{\textwidth}{!}{\begin{tikzpicture}[-,auto]
\begin{axis}[
		view={0}{90},
		point meta min=0, 
		point meta max=1,
		colormap={yellowblue}{color=(yellow) color=(blue)},  
		-,
 		xmax=1,
		ylabel=Normalized Profit,
		xlabel=Normalized Capacity,
	    grid = both,
 		major grid style={line width=.5pt,draw=black!10},
 		minor grid style={line width=0.1pt,draw=black!10},
	    major tick length=0pt,
	    minor tick length=0pt,
	    minor tick num=0,
 	    legend pos=south east,
 	    height=6cm,
 	    width=7cm
	 	]
		 	
	\addplot3[
 		mesh/cols=25,	
 		mesh/rows=25,	
		contour gnuplot={
			levels={0.4,0.6}, 
			draw color=mapped color, 
			label distance=100pt,
			contour label style = {
				    /pgf/number format/fixed,
				    /pgf/number format/precision=2
			}
		}
 	] table [col sep=comma] {actual_bounds-50.csv};\addlegendentry{Actual}

	\addplot3[
 		mesh/cols=21,	
 		mesh/rows=21,	
		contour gnuplot={
			levels={0.4,0.6}, 
			draw color=mapped color,
			label distance=100pt,
			contour label style = {
				    /pgf/number format/fixed,
				    /pgf/number format/precision=2
			}
		},dashed 
 	] table [col sep=comma] {lower_bounds-50.csv};\addlegendentry{Lower bound}

\end{axis}

\end{tikzpicture}}
     \caption{\small{Isoquants: Actual vs Lower bound.}}\label{fig:bounds}
\end{subfigure}
\end{tabular}
\caption
{\small Phase transition and time complexity for knapsack instances with 50 items, weights/values sampled from a uniform distribution from a range of $[0,10^7]$.}
\label{fig:phase-transition}
\end{figure*}

%

%
%


\section{Computational experiments}\label{sec:results}

We now empirically study the relationship between the time complexity of solving random KP instances and the phase transition in solvability; and the tightness between the actual probabilities and the theoretical lower bounds.
We will consider random instances with $n \in \{ 20, 30, 40, 50 \}$ items.
For each $n$, we sampled (with replacement) a collection of weight and value combinations (a combination is $(\tup{\hat{w}_1, \ldots, \hat{w}_n}, \tup{\hat{v}_1, \ldots, \hat{v}_n})$). 
Weights and values were always sampled from the same discrete uniform distribution with range $[0,10^7]$. 
In the second step, for each combination, multiple knapsack instances were generated by considering normalized capacity and normalized profit constraints at regular intervals of $0.04$ in $[0, 1]$, respectively.


To solve the instances, we used two existing generic off-the-shelf solvers that are based on different solving methods, \emph{Minizinc}~\cite{minizinc} and \emph{Minisat+}~\cite{minisatp}. Minizinc is a constraint solver that uses a branch-and-bound technique with constraint propagation. Minisat+, on the other hand, is a pseudo-boolean satisfiability solver (based on Minisat) that uses conflict resolution \cite{DPLL}. 
We used on the ``knapsack'' constraint of Minizinc library to use a state-of-the-art KP solver.
We used these two different solvers to \replaced{reduce}{exclude} the possibility that our results are biased by a particular solution technique.
The results of the computational experiments for instances with 50 items are shown in Figure~\ref{fig:phase-transition} (results for instances with different numbers of items were qualitatively different). 

\paragraph{Solvability}
First, we examine whether the probability of solvability of instances changes as a function of normalized capacity $c$ and normalized profit $p$ and, in particular, whether this probability exhibits a phase transition. 
The plot in Figure~\ref{fig:solve:c-p} shows the probability  that an instance had a solution at given levels of normalised capacity $c$ and normalised profit $p$.  As is evident from the plot, the probability that an instance has a solution tends to increase in $c$ and decrease in $p$, ceteris paribus, as expected. In other words, solvable instances generally have relaxed constraints (i.e., normalised capacity is higher than normalised profit) while unsolvable instances have tight constraints (i.e., normalised capacity is lower than normalised profit).

In a contour plot of probability of a solution as a function of $c$ and $p$ with equi-distant contours, is be \emph{above} the $45$ degree line (see Figure~\ref{fig:bounds}). 
Likewise, in a plot of $P[E(r)]$ as a function of the log-ratio $\log (r) = \log (c/p)$, the region of maximum uncertainty---where the probability of a solution is $1/2$---will be to the \emph{left} of $0$ (see Figure~\ref{fig:solve:ratio}). 
\paragraph{Time complexity}
Next, we examine time complexity of the instances.
We computed the time taken to solve the random instances by Minizinc (Figure~\ref{fig:stime-mzn} and \ref{fig:solve:ratio}). 
%
%
The plots show that the hard instances occur around the phase transition from solvability to unsolvability (region of maximum entropy in the probability space). 
%
%
%
The average compute time using Minisat+ (not shown here) had same relationship to phase transition as Minizinc.
Hard instances for both solvers were located in the same area of the parameter space, indicating robustness to our findings with regards to the solution technique. 
Also observe that we are able to identify an island of difficulty within the instances that lie in the phase transition (see Figure~\ref{fig:stime-mzn}).
The hardest instances occur between normalised capacity of 0.4 and 0.52 and normalised profit of 0.52 and 0.68, respectively. 
\paragraph{Phase transition}
To characterise the phase transition in the knapsack problem in terms of $c$ and $p$, we plot the probability of solvability as a function of $c$ and $p$ (Figure~\ref{fig:solve:c-p}). 
We also use the log of the ratio (i.e., $\log(c/p)$, to scale the x-axis) to plot the solvability against one metric (Figure~\ref{fig:solve:ratio}). 
Note that similar to phase transitions observed in other NP-complete problems such as 3-SAT, integer partition and graph colouring, the phase transition from solvability to unsolvability occurs precipitously near $0$, that is, near the point $c=p$. 

Importantly, the probability that an instance has a solution, has a \emph{phase transition above the 45 degree line in the (c,p) space}. 
Our computational results also show that the transition from solvability to unsolvability exhibits a {\it convex} shape. In the area near the centre of the diagonal where $c=p$, instances with $c < p$ have a higher probability of having a solution at the centre of the diagonal than instances at the ends of the diagonal.
This finding confirms our result in Section~\ref{sec:formal}.

 This convexity is related to structural properties of the search space. Intuitively, the number of possible knapsack configurations is maximal when the size of the knapsack is half the total number of items (the function $\binom{n}{k}$ peaks at $k=n/2$). In addition, given that we are considering random instances, there is a probability that there are some items whose weights are (relatively) smaller than their values. Therefore, the actual probability of finding a knapsack configuration that is a solution to an instance with lower normalised capacity than normalised profit, peaks when the knapsack configuration size is half of the total number of items.

With the aid of simulations, we now investigate how tight the lower bound is. We randomly generate 1,000 knapsack instances with $n = 50$, normalize weights and values, apply the weight-greedy strategy and determine, for each combination of $c$ and $p$ (in intervals of 0.05, from 0 to 1), whether there exists a solution. Figure~\ref{fig:bounds} plots the isoquants in (c,p) space. In $c-p$ space, the phase transition region clearly shows convexity; virtually all (equi-distant) isoquants are above the 45 degree line. The region exhibits the same features as the one obtained from the true (estimated) probabilities. 


\section{Human Experiments}\label{sec:human-experiments}

As argued before, the knapsack problem is ubiquitous in every-day human life, at many different levels of cognition \cite{Torralva:2013ADHD}. An interesting question, therefore, is whether instances near the phase transition are also harder for humans.
At the surface, there are many inherent differences between electronic and human computers~\cite{MurawskiBossaerts:SR16}. 
For example, as compared to electronic computers, humans are more memory constrained and therefore have limited capacity to implement solution techniques such as dynamic programming. In addition, humans are affected by relatively short attention spans, fatigue, and calculation errors. On the other hand, recent experimental evidence suggests that instance complexity does predict human behaviour. It has been shown that as instance complexity of the optimisation variant of the knapsack problem increases, the probability of human participants being able to solve the instance decreases~\cite{Meloso:2009go,MurawskiBossaerts:SR16}.

In the following, we examine whether instances near the phase transition identified above are also harder for humans. To investigate this question, we used  data from an experimental study where human participants were asked to solve a number of instances of the optimisation variant of the 0-1 knapsack problem.
We take the data from the optimisation task. 
Twenty-two human volunteers (age range = 18-30, mean age = 22.2, 17 female, 5 male) recruited from the general population took part in the study. 
The experimental protocol was approved by our university’s human research ethics committee. 
%
The order of instances was randomised in each session. 
Participants were incentivized as follows: (i) for each instance they received a cash amount proportional to the total value of their knapsack (relative to the value of the optimal knapsack); and (ii) a fixed show-up fee. The experiment included a training session prior to actual testing. 
It can be considered as a sequence of decision tasks in which participants have to answer the question ``Does there exist another set of items with a higher profit than the current set?''. We take the time spent at each node as a proxy for the time participants spent on solving the corresponding decision problem.
 In the experiments, participants were asked to find the set of items with the highest total value subject to the capacity constrained. They always started with an empty knapsack and had 240 seconds to solve an instance. 
 They used a computer interface to add items to and remove items from the the knapsack. 
 Each participant solved 15 instances. Each instance had 10 items. Item values and weights, as well as capacity, differed across instances. Values and weights for the different instances were drawn from the same distribution. 

\begin{figure}[t]
\begin{tabular}{lr}
\begin{subfigure}[b]{0.4\textwidth}
   \begin{tikzpicture} 
\tikzstyle{every picture}=[-]
	\begin{axis}[
  		width=\textwidth,
		xlabel={$log({c}/{p})$}, 
		ylabel={Average Time (s)},
		label style={sloped},
		ymax = 120,
		ytick = {0,20,40,60,80,100,120},
		xmin=-1,
		grid = both,
 		major grid style={dotted,line width=.5pt,draw=black!40},
 		minor grid style={line width=0.1pt,draw=black!10},
	    major tick length=0pt,
	    minor tick length=0pt,
	    minor tick num=0,
	    axis line style = {},
	    ylabel near ticks
	]
	 \addplot+ [
	 	only marks, 
	 	color=blue,
	 	mark=+,
	 	mark options=solid, mark size=1pt
	 ]
		 table[x=x, y=time, col sep=comma] {non_terminal_mean.csv};
	 \addplot+ [
	 	only marks, 
	 	color=red,mark=*,
	 	mark options=solid, 
	 	mark size=1pt
	 ]
		 table[x=x, y=time, col sep=comma] {terminal_mean.csv};
	\end{axis}
\end{tikzpicture}
    \caption{\small{Average time spent on nodes.}}\label{fig:human:c-p}
\end{subfigure}
&
\begin{subfigure}[b]{0.4\textwidth}
     \begin{tikzpicture}
\tikzstyle{every picture}=[-]
	\begin{axis}[
  		width=\textwidth,
		xlabel={$log({c}/{p})$}, 
		ylabel={Average Time (ms)},
		label style={sloped},
		xmin=-1.5,
		xtick={-1,0,1,2,3},
		grid = both,
 		major grid style={dotted,line width=.5pt,draw=black!40},
 		minor grid style={line width=0.1pt,draw=black!10},
	    major tick length=0pt,
	    minor tick length=0pt,
	    minor tick num=0,
	    ylabel near ticks
	]
	 \addplot+ [
	 	only marks, 
	 	color=black,
	 	mark=+,
	 	mark options=solid, mark size=1pt
	 ]
		 table[x=x, y=time, col sep=comma] {mzn_c_p_time.csv};

	\end{axis}
\end{tikzpicture}
     \caption{\small{Average time by Minizinc}}\label{fig:time-mzn}
\end{subfigure}
\end{tabular}
  \caption{Human complexity of solving Knapsack instances.}
  \label{fig:human-solvers}
\end{figure}
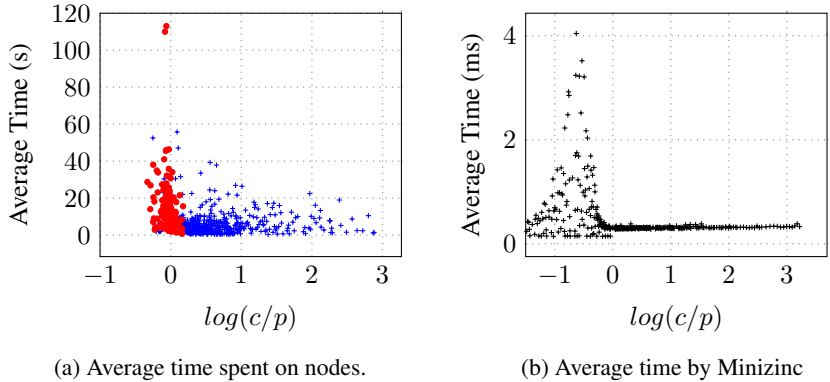

We call each addition of an item to and removal from the knapsack a \emph{move}. We measured the time taken between each move. We then calculated normalised capacity $c$ and normalised profit $p$ for each feasible search state that participants visited (participants could not reach infeasible search states, that is, states that violated the capacity constraint).

Figure~\ref{fig:human-solvers} shows two plots that compare the average time spent by human participants and by Minizinc solver (on instances with 20 items) as a function of $log(c/p)$.
%
The nodes colored in red are the nodes corresponding to local maxima in the search space (i.e., states in which no additional item could be added without violating the capacity constraint).

As Figure~\ref{fig:human:c-p} shows, the average time taken is maximal around the vicinity of the phase transition (i.e., just before $log(c/p)=0$). Note that the time spent by participants in a given state is a function of two key cognitive processes: (i) parsing of information (profit and weight of the state); and (ii) solving of instance of the decision problem (does there exist a feasible state with a higher total profit?). At the start of each instance, a participant will consume more time to parse  information (e.g., weight and value combinations of items) and therefore the time absorbed by this process will be higher.
After having parsed the information, the time taken to solve an instance would be the main driver of the time spent in a state.
We argue that the high time spent on the nodes that have high $log(c/p)$ values is because of the time taken to parse the instance.
As is evident from Figure~\ref{fig:human-solvers}, time increases the closer a search node is to the phase transition boundary. This finding provides initial evidence that instances near the phase transition were also harder for humans.


\section{Discussion}\label{sec:discussion}

Problems in the class NP-complete are considered intractable if, in the worst case, the time taken by a computer to solve them grows faster than polynomially in input size. However, many instances of NP-complete problems can nevertheless be solved in polynomial time. 
Understanding what the hard instances are and how they affect human decision making, is an important question. 

\citet{Cheeseman:PT1991} conjectured that all NP-complete problems have at least one order parameter such that the phase transition lies on a critical value of this parameter. NP-complete problems for which such parameters have been identified include satisfiability~\cite{MitchellSelmanLevesque:AAAI92-PhaseTransition,GentWalsh:AIJ94}, integer partition~\cite{Gent:CI1998}, graph colouring~\cite{Cheeseman:PT1991}, and traveling salesman~\cite{Gent:TSP1996}.
In this paper, we identified such a parameter for the 0-1 KP providing further evidence towards the conjecture.
 
A generalization that links problems with phase transitions is based on the notion of constrainedness of  instances~\cite{Gent:ConstrainednessOfSearch}. Constrainedness of search over an ensemble can be estimated by the formula $ \kappa = 1 - \frac{log_2(\ept{[Sol]})}{N}$ , where $N=log_2(|S|)$ (i.e., log of the size of the total search space $S$), and $\ept[Sol]$ is the expected number of solutions. Instances with $\kappa<1$ are considered under-constrained while instances with $\kappa>1$ are over-constrained. Phase transitions have been shown to exist (e.g., for 3-SAT and graph colouring) when $\kappa\approx 1$. We showed that for the knapsack problem (under the sampling assumptions of Section~\ref{sec:formal}), $\kappa\approx 1$ when $c/p \approx 1$, where $c$ and $p$ are normalized capacity and normalised profit, respectively.


 In this paper, we also  provide initial evidence  that difficulty for human decision-makers increases closer to the phase transition.
We analysed data from an optimisation task by looking at the average time spent on each search node.
%
%
 %
 In future work, we want to investigate how human reasoning is affected by phase transition in the decision variants of the knapsack problem, and the relation between its decision and optimisation variants.
%


%
\small
\bibliographystyle{plainnat}
\bibliography{knapsack}

\end{document}